\documentclass[runningheads]{llncs}

\usepackage{graphicx}
\usepackage{times}
\usepackage{latexsym}
\usepackage{url}
\usepackage{times}  
\usepackage{helvet}  
\usepackage{courier}  
\usepackage{url}  
\usepackage{graphicx}  
\usepackage{latexsym}
\usepackage{graphicx}
\usepackage[numbers, sectionbib]{natbib}
\usepackage{amsmath,amssymb}
\usepackage{caption}
\usepackage{courier}
\usepackage{epstopdf}
\usepackage{helvet}
\usepackage{url}
\usepackage{multirow}
\usepackage{color}
\usepackage{lipsum}

\newcommand{\tabincell}[2]{\begin{tabular}{@{}#1@{}}#2\end{tabular}}

\begin{document}
\title{Domain Representation for Knowledge Graph Embedding}
%
%

\author{Cunxiang Wang\textsuperscript{1,2}, Feiliang Ren\textsuperscript{3}, Zhichao Lin\textsuperscript{3}, Chenxu Zhao\textsuperscript{3}, Tian Xie\textsuperscript{3} and Yue Zhang\textsuperscript{2}\\
\vspace{-5pt}
\institute{College of Computer Science and Technology, Zhejiang University, China \email{wangcunxiang@westlake.edu.cn} \and School of Engineering, Westlake University, China \\ \email{yue.zhang@wias.org.cn} \and School of Computer Science and Engineering, Northeastern University, China
\\ \email{ renfeiliang@cse.neu.edu.cn, enjoymath2016@163.com
\{ch4osmy7h,thankoder\}@gmail.com}}}
%
%

\maketitle              

\vspace*{-1cm}
\begin{abstract}
    Embedding entities and relations into a continuous multi-dimensional vector space have become the dominant method for knowledge graph embedding in representation learning. However, most existing models ignore to represent hierarchical knowledge, such as the similarities and dissimilarities of entities in one domain. We proposed to learn a Domain Representations over existing knowledge graph embedding models, such that entities that have similar attributes are organized into the same domain. Such hierarchical knowledge of domains can give further evidence in link prediction. Experimental results show that domain embeddings give a significant improvement over the most recent state-of-art baseline knowledge graph embedding models. 
\keywords{Representation learning  \and Knowledge graph \and Domain}
\end{abstract}
\section{Introduction}
Containing relational knowledge between entities, Knowledge Graphs \cite{NTN, TransR,STransE, InforSeek_Response_Ranking} can help improve reasoning in QA systems \cite{KAANN}, conversation systems \cite{InforSeek_Response_Ranking} and recommendation systems \cite{Huang:2018:ISR:3209978.3210017}. Intuitively, a more comprehensive knowledge graph will be more beneficial for its applications. But knowledge graphs are far from complete \cite{NTN}. Knowledge graph embedding models can be helpful for expanding knowledge graphs. The basic idea is to project entities and relations into a continuous multi-dimension space so that new relational facts can be scored for their credibility in a dense vector space. This task is called link prediction \cite{transE}. 

Numerous models have been proposed for knowledge graph embedding, including TransE \cite{transE}, and TransR \cite{TransR}, which learn embeddings of entities and relations by leveraging their distributed contexts. One issue of TransE and its subsequent variants, however, is that knowledge is not organized hierarchically. For example, all geographic knowledge share common attributes, a categorical representation of which can facilitate link prediction. We address this issue by introducing the concept of \emph{domains}, which are collections of entities around a certain relation. For example, consider the relation ``Capital'', for which the head entity must be a country, and the tail entity must be a capital city. In this example, the set  ``country'' and the set ``capital city'' are both \emph{domains}.

Domains add a layer of abstraction to KG embeddings. Knowledge on whether a given entity belongs to a given relation's head or tail domain is helpful for link prediction. For example, when considering two triples (USA, capital, Washington) and (USA, capital, New York) as candidate new triples, Washington D.C. can be inside the ``capital city'' domain while New York can be outside the domain. As a result, (USA, capital, New York) should suffer a penalty given such domain knowledge. In contrast, preliminary experiments show that when using TransR/STransE for link prediction, more than half of incorrect entities do not belong to the right domain.

We propose a model for learning explicit domain representation given a KG. Since entities in a domain are similar in some attributes, while dissimilar in other attributes, domains are restricted using hyper-ellipsoids in the vector space. Given a trained KG, we learn the representation of each domain by fitting one hyper-ellipsoid to a hyper-point cluster with similar attributes. The training objective is set to minimize the overall distance between entities and the domain surface. We approximately calculate the distance measure for a simple, concise algorithm with high runtime efficiency. The underlying KG embedding models or their embeddings do not change during training. In link prediction, we calculate the distance between each candidate entity and the target domain, adding the distance to the baseline score to rank candidate entities.

Experiments show that our model is effective over strong baseline models. To our knowledge, we are the first to explicitly learn domain representations over knowledge graph embeddings. We release our source code and models at https://github.com/wangc
\
unxiang/Domain-Representation-for-Knowledge-Graph-Embedding.

\section{Related Work}

Prior knowledge graph embedding models and the works about domains are related to our model. Knowledge graph embedding models are either employed with external information or not. For those without external information, there are two main streams-{\it Translation-based models} and {\it neural network models}. Models using external information use different types of resources. 

\subsection{Models without Using External Information}
{\it Translation-based models} treat entities as a hyper-points and relations as a vectors in the vector space. The training objectives are set to ensure certain correlations between points and vectors. TransE \cite{transE} is a seminal work of all translation-based models, it believes that head entity plus relation approximately equals tail entity in the vector space. 
Subsequently, TransH \cite{transH} overcomes the flaws of TransE concerning
the 1-to-N/N-to-1/N-to-N relations. 
TransR \cite{TransR} builds entity and relation embeddings in separate entity and relation spaces. 
TransSparse \cite{TranSparse} aims to handle heterogeneity and imbalance of data, and STransE \cite{STransE} models head and tail spaces differently.
TransD \cite{transD} considers entities for projection matrices. TransA \cite{TransA} makes the margin changes dynamically. TransG \cite{TransG} aims to solve the problem of multiple-relation semantics, and ITransF \cite{ItransF} uses sparse attention to solve the problem of data sparsity.

Among {\it neural network models}, SLM (Single Layer Model) \cite{NTN} applies the neural network to knowledge graph embedding. 
NTN (Neural Tensor Network) \cite{NTN} uses a bilinear tensor operator to represent each relation.
ProjE \cite{ProjE} can be seen as a modified version of NTN.

We choose translation-based models as baselines since they are more efficient and highly effective compared to neural network based models. 

\vspace{-8pt}
\subsection{Models Using External Information}
{\it Text-aware models} import external information. The main idea is to employ textual representation or attributes information of entities and relations to existing models (e.g. TransE), which also means that text aware models cannot work independently and have to be attached to a knowledge graph embedding model to improve the baseline model performance. In this sense, text-aware models are similar to our domain representation model. 
However, our model does not need any external information. 


\subsection{Investigation of Domains}
Some research on knowledge graphs can be regarded as domain related. 
For example, Dual-Space Model \cite{domain} is designed for calculating the similarity between different domains and functions to help semantic similarity task, but cannot be used for link prediction. Another example, \cite{Yang} utilize learned relation embeddings to mine logic rules, such as {\it BornInCity(a,b)} \^ {\it CityOfCountry(b,c)}  $\Rightarrow$ {\it Nationality(a,c)}. The concept of domains is used to restrict search choices of logic rules. Though the models above use entities in domains, none of them tries to represent domains explicitly in the vector space, which is the core idea of our model. As a result, they cannot extract the common attributes of domains.

\section{Baselines}

In this section, we introduce the baseline models - {\it Translation-based models} including TransE \cite{transE}, TransR \cite{TransR} and STransE \cite{STransE}, as well as the main task of link prediction.

\subsection{Three Translation-based models} 

TransE \cite{transE} is the root of all translation-based models. As a seminal work, given a head entity \(h\), a relation \(r\) and a tail entity \(t\), TransE \cite{transE} models a relation triple\(\  \langle h,r,t \rangle \ \) with \(h + r \approx t\). The training objective function is thus to minimize 
\vspace*{-0.5\baselineskip}
\begin{equation}
f_{r}\left( h,t \right) = \left\| h + r - t \right\|_{l_{1/2}}\
\vspace*{-0.5\baselineskip}
\end{equation}

over a whole KG. Pre-trained with the head entity vectors \(h\)s, the relation vectors \(r\)s and tail entity vectors \(t\)s by TransE, TransR \cite{TransR} uses one projection matrix per relation to do translational operation in the relation space, with an objective function 
\vspace*{-0.5\baselineskip}
\begin{equation}
f_{r}\left( h,t \right) = \left\|  W_{r}h + r - W_{r}t \right\|_{l_{1/2}}\
\vspace*{-0.5\baselineskip}
\end{equation}
where \(W_{r} \in R^{k \times d}\) is the projection matrix. 
STransE \cite{STransE} is similar to TransR; It also uses pre-trained entity vectors and relation vectors outputted by TransE. But for STransE, each relation has two projection matrices, one for head entities, the other for tail entities. STransE's objective function is
\vspace*{-0.5\baselineskip} 
\begin{equation}
f_{r}(h,t) = \left\| W_{r,1}h + r - W_{r,2}t \right\|_{l_{1/2}}\
\vspace*{-0.5\baselineskip}
\end{equation}
where \(W_{r,1} \in R^{k \times d}\) and \(W_{r,1} \in R^{k \times d}\) are the projection matrices for head entities and tail entities, respectively.

\subsection{Link Prediction}

{\it Link prediction} is one of the most common evaluation protocols in knowledge graph embedding. Given a test triple  \(< h,r,t >\), where \(h\), \(r\) and \(t\) denote the head entity, the relation and the tail entity, respectively. The task is to predict the second entity (\(h\) or \(t\)) once the relation and one entity are determined. We first remove the head entity, replacing it with all entities in the knowledge graph to calculated a fitness score for each entity, based on the objective function of the current model. All the entities are ranked according to the score, and the rank of \(h\) among all entities is recorded. We repeat the same procedure for the tail entity. The two ranks are used for the credibility score of the triple \(< h,r,t >\), which will be used in subsequent evaluation metrics.

\section{Domain Representation Using Ellipsoids (DRE)}

We model domain structures in the vector space as hyper-Ellipsoids. Section 3.1 introduces a formal definition of domains and explains why we use hyper-ellipsoids to represent domains. Section 3.2 discusses the formal representation of a hyper-dimensional hyper-ellipsoid. Section 3.3 describes how domains can be used for tasks related to knowledge graphs. Section 3.4 discussed training. Finally, Section 3.5 illustrates how the domain model can be used in combination with various knowledge graph embedding models.

\subsection{Domain and Hyper-Ellipsoid}
We define a \emph{domain} in a knowledge graph as the set of a relation's head or tail entities. Formally, for any relation \(r\), its head domain \(D_{h,r}\) is 
\vspace*{-0.5\baselineskip}
\begin{equation} 
D_{h,r}= \{e_{h} \in E |\ \exists e_{t} \in E \land (e_{h}, r, e_{t}) \in T\} 
\vspace*{-0.5\baselineskip}
\end{equation}
And its tail domain \(D_{t,r}\) is 
\vspace*{-0.5\baselineskip}
\begin{equation} 
D_{t,r}= \{e_{t} \in E |\ \exists e_{h} \in E \land (e_{h}, r, e_{t}) \in T\} 
\vspace*{-0.5\baselineskip}
\end{equation}
where \(T\) is the set of all triples, \(E\) is the set of all entities.

From our perspective, the concept of a domain is closely related to the concept of similarity. Entities in a domain have similar attributes and are close in certain dimensions in the vector space. Other the other hand, the shape of a domain in a vector space distributes unevenly in different dimensions since the head/tail entities are only similar in some attributes, but unrelated or even distinct in others. For instance, to the relation ``Capital'', ``Beijing'' and ``Washington''  can both be tail entities. However, they can be very different in other senses. For example, ``Beijing'' is a large city with more than 20M people while ``Washington'' is a small city with only 700K people, and they are also far away from each other in geological location. Thus in the embedding space, the dimensions describing the ``capital'' attributes will be close, but those describing population and location attributes will be distant.
\begin{figure}{}
  \centering 
  \includegraphics[width=2.01011in,height=1.88993in]{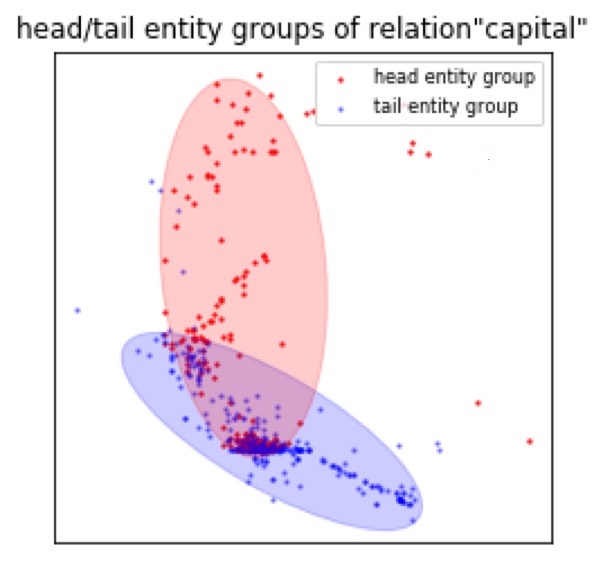}
  \caption{One relation and its head/tail entity group, trained on FB15K by TransE (vectors not normed). Most head/tail entities are close in one dimension but scattered in other dimensions respectively.} 
  \vspace*{-15pt}
\end{figure}

Therefore, the space of a domain reflected in the vector space can be an enclosure with narrow boundaries in some dimensions and with medium or wide boundaries in others. Figure 1 illustrates the observation. This fits the shape of a hyper-ellipsoid, which can rotate freely.

\subsection{Ellipsoid in Hyper-Dimensional Space}
 A hyper-ellipsoid can be represented as:
 \vspace*{-0.5\baselineskip}
\begin{equation}
  (x-a)^TM(x-a)=1
  \vspace*{-0.5\baselineskip}
\end{equation}

\noindent
where
\(x,a \in R^{n}\), and \(M \in R^{n*n}\). \(M\) should be a positive definite matrix. According to Cholesky decomposition,
\vspace*{-0.5\baselineskip}
\begin{equation}
  M=LL^T
  \vspace*{-0.5\baselineskip}
\end{equation} 

\noindent
where \(L\) should be a lower triangular matrix. In the training period, we decompose M to L, and update L instead of M. L will still be a lower triangular matrix in training. Using L to calculate M, we can assure that M remains a Hermitian matrix in the training period.

\subsection{Distance Between Entity Vectors and Hyper-Ellipsoid}

We use an approximate method to calculate the distance between a point and the surface of an ellipsoid in the vector space. First, a straight line is used to join the entity point with the geometric center of the hyper-ellipsoid. Then we work out the crossing point of the straight line and the hyper-ellipsoid’s surface. Finally, the distance between the crossing point and the entity point is calculated, which is our defined distance between the entity vector and the hyper-ellipsoid. Figure 2 illustrates the distance.

\vspace*{-0.5\baselineskip}
\begin{figure}{}
  \centering 
  \includegraphics[width=2.01011in,height=1.88993in]{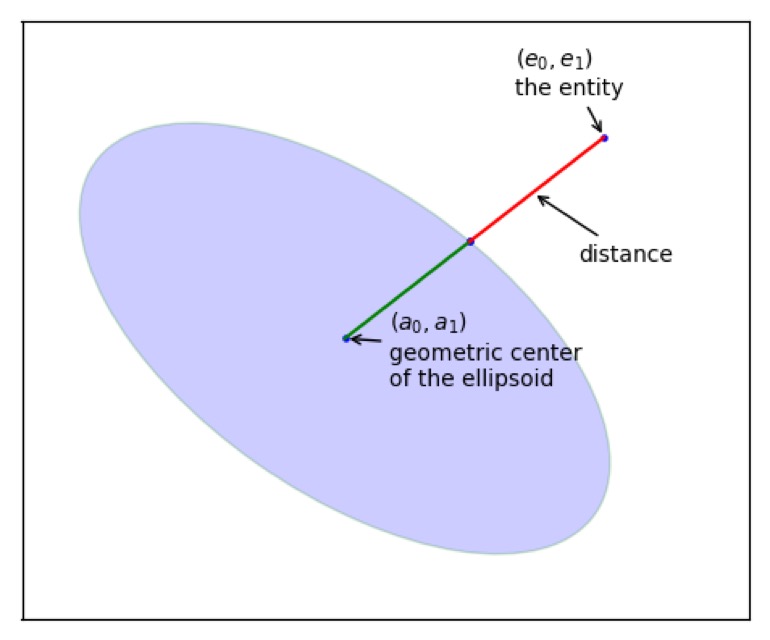}
  \caption{An entity point and the surface of an ellipsoid, the red line is our defined distance.} 
  \vspace*{-15pt}
\end{figure}

The distance between the entity vector and the surface of the ellipsoid is:
\vspace*{-0.5\baselineskip}
\begin{equation}
  D = \left| 1 - \frac{1}{\sqrt{\left( e - a \right)^{T}M\left( e - a \right)}} \right|\left\| e - a \right\|_{2}
  \vspace*{-0.5\baselineskip}
\end{equation} where \(e,a \in R^{n}\), and \(M \in R^{n*n}\), \emph{e} is a vector representing the entity, and \(a\) is a vector representation of the hyper-ellipsoid's geometric centre.

Note that the use an approximate distance is mainly because the exact distance is complex with many constraints, which can be infeasibly slow to calculate.

\subsection{Score Functions}

The fitness between an entity vector and a domain is measured according to the distance \(D\) in equation 8. The lower the score, the better the entity fits the domain. In the training process, to fit the hyper-ellipsoid to a domain, we set the score function as:
\vspace*{-0.5\baselineskip}
\begin{equation}
f_{{train}}(e,E)=D
\vspace*{-0.5\baselineskip}
\end{equation} where \(e\) indicates an entity, \(E\) indicates an Ellipsoid and \(D\) indicates the geometric distance discussed above. Note that all \(e\)s have been readily trained an a knowledge graph, and they will not change in ellipsoid-training. Only \(M\)s and \(a\)s are trained.

In the testing process, if an entity is inside the hyper-ellipsoid, it belongs to the domain, and we set \(f\) to 0; if an entity is outside the hyper-ellipsoid, it still can belong to the domain. We use \(D\) to describe the relatedness between the entity and the domain, setting \(f = D\).

Thus, the score function in testing is: 
\begin{align} 
f_{test}(e,E) =\begin{cases} 0 
& (e-a)^TM(e-a) < 1 \\ D & (e-a)^TM(e-a)\geq1\end{cases} \end{align} 
where $(e-a)^TM(e-a)<1$  means the entity is inside the hyper-ellipsoid, \\\(\left( e - a \right)^{T}M\left( e - a \right) \geq 1\) means the entity is outside the hyper-ellipsoid, and \(D\) indicates the geometric distance.

The training score ensures that the hyper-ellipsoid will fit the space of the domains and the testing score ensures that all entities belonging to a certain domain are treated equally, but the entities outside the domain are estimated by their distance from the domain.

\vspace{-12pt}
\subsection{Training}
Our training goal is to find a hyper-ellipsoid that best fits a cluster of the entities that represents a domain. The training objective is:
\vspace*{-0.5\baselineskip}
\begin{equation}
  L = \sum_{e \in Dom}^{}{\min\left( f_{\text{train}}\left( e,E \right) \right)}
  \vspace*{-0.5\baselineskip}
\end{equation} 
where \(e\) is an entity belonging to a domain \(Dom\), and \(min()\) aims to minimize the distance between the hyper-ellipsoid and the entities s. We use SGD to minimize \(f_{\text{train}}\) by updating the ellipsoid parameters.

For models having multiple spaces, we only choose the embeddings in the ``final'' space to train. For example, for TransR, we choose the projected space (relation space). So the model does not care about how different spaces are projected.

\vspace*{-5pt}
\begin{figure}[t]
  \centering 
  \includegraphics[width=3.01011in,height=1.88993in]{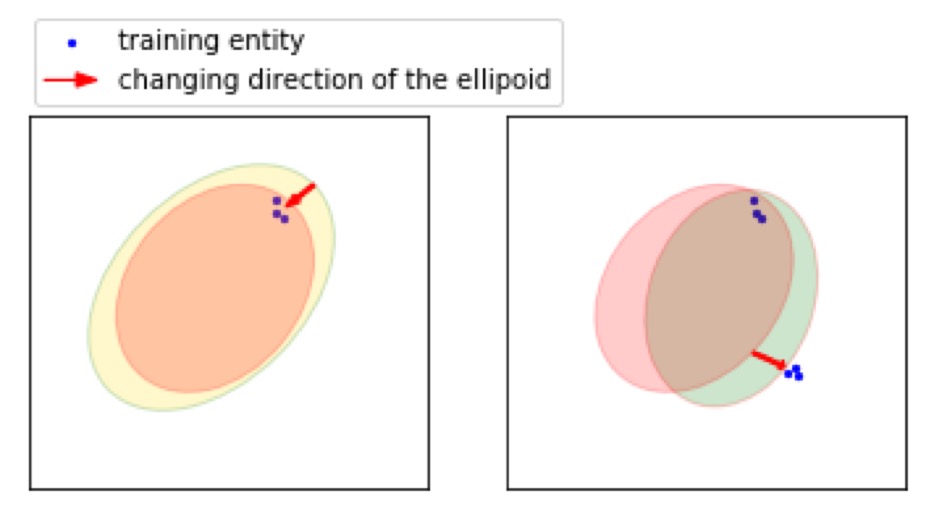}
  \caption{Simulated training process. A hyper-ellipsoid changes its size, shape and orientation during training. The left figure shows that when training entities are inside the hyper-ellipsoid, boundaries of the ellipsoid will move inward. The right figure shows that when training entities are out of the hyper-ellipsoid, boundaries of the ellipsoid will move outward. } 
  \vspace*{-15pt}
\end{figure}

\section{Combination with KG Embedding Models}
Domain representations are combined with baseline knowledge graph embedding models to enhance the power of entity distinguish. When doing link prediction, entities not belonging to a domain receives a penalty over the baseline model scores based on their spatial distance from the domain. However, entities belonging to the domain do not receive such penalty scores.

In the testing process, we add the score function of DRE to the baseline models:
\vspace*{-0.5\baselineskip}
\begin{equation}
  f_{r}\left( h,t \right) = f_{\text{test(baseline)}}\left( h,t \right) + f_{\text{test(DRE)}}\left( e,E \right)
  \vspace*{-0.5\baselineskip}
\end{equation} 
where \(f_{\text{test(baseline)}}\left( h,t \right)\) is the score function of the baseline model, and \(f_{\text{test(DRE)}}(e,E)\) is the penalty score of our model.

\begin{figure}[t]
  \centering 
  \includegraphics[width=2.01011in,height=1.88993in]{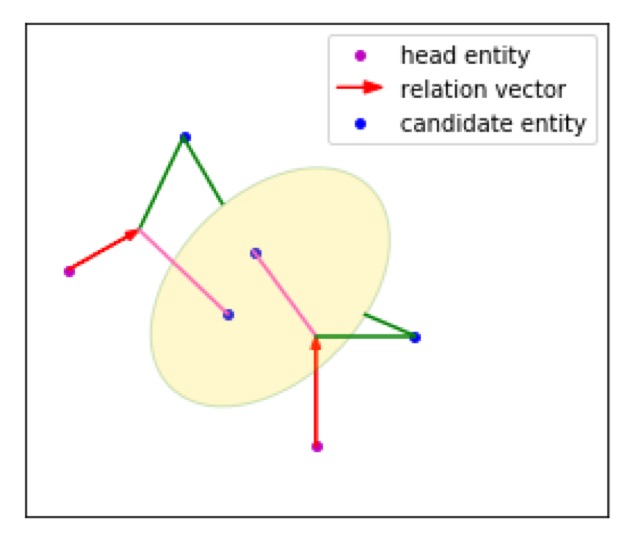}
  \caption{The testing process of a translation-based model combined with DRE. For candidate entities inside the ellipsoidy, their scores have no change from the baseline models, which are indicated by the pink lines. However, for candidate entities outside the ellipsoid (outside the domain), their scores are the original scores augmented by the scores of DRE (the distance between the candidate entities to the ellipsoid’s surface), which are the two intersecting green lines.  } 
  \vspace*{-15pt}
\end{figure}

\subsection{Example}
\vspace{-2pt}
We use STransE as an example to illustrate how our model DRE is applied on top of a knowledge graph embedding model. STransE is one of the best performing translation-based models, and its score function is \(f_{r}(h,t) = \left\| W_{r,1}h + r - W_{r,2}t \right\|_{l_{1/2}}\), where \(W_{r,1},\ W_{r,2}t\) are the projection matrices. 

The process consists of five steps:
\vspace{-5pt}
\begin{enumerate}
\def\labelenumi{\arabic{enumi}.}
\item
  Train a TransE model to obtain entity embeddings \(h\)s, \(t\)s and relation embeddings \(r\)s, mentioned in Equation 1.
\item
  Use the TransE embeddings to train STransE and obtain new entity and relation embeddings as well as the projection matrices, shown in Equation 3.
\item
  Use the entity embeddings and matrices from STransE (in the projected space) to train hyper-ellipsoids. Given a triple\(< h,r,t >\), The projected embeddings of the two entities are used to train two domains - the head doamin and tail domain of the relation \(r\).
\item
  For each testing case, calculate
  \(f_{\text{test(STransE)}}\)
  and \(f_{\text{test(DRE)}}\) respectively and add them for a final score \(f = f_{\text{test(STransE)}} + f_{\text{test(DRE)}}\). 
\item 
  Use \(f\) in link prediction.
\end{enumerate}

\vspace{-15pt}
\section{Experiments}
\vspace{-7pt}
We for comparing our domain representation model with the baseline models on link prediction, evaluating the results using mean rank, Hits@10/3/1 of TransE/TransR/STransE combined with DRE, respectively.
\vspace{-15pt}
\subsection{Datasets}
We conduct our experiments on two typical knowledge graphs, namely WordNet \cite{WordNet} and Freebase \cite{Bollacker2008Freebase}, choosing the dataset WN18 from Wordnet and the dataset FB15K from Freebase. \footnotemark[1] Information of these two databases is given in Table 1.

\footnotetext[1]{Because of reverse relations, FB15k-237 and WN18RR have attracted much attention. However, DRE is free from this problem because each domain is independent. Besides, more papers reported results on WN18 and FB15K than those on FB15k-237 and WN18RR.So we choose the more general and widely-used datasets WN18 and FB15K.}
\vspace{-15pt}
\subsection{Evaluation}

Based on link prediction, two measures are used as our evaluation metric. \textbf{Mean Rank}: the mean rank of correct entities in all test triples. \textbf{Hits@10/3/1}: the proportion of correct entities ranked at top 10/3/1 among all the candidate entities.
 
Following \cite{transE} and \cite{TransR}, we divide over evaluation into \textbf{Raw} and \textbf{Filtered}. Some triples obtained by randomly replacing entities in gold triples are also correct, such triples exist in the training, validation or test sets. If we do not remove the corrupted triples when calculating the rank of correct triples, the evaluation method is called \textbf{Raw}; if we remove them, the evaluation method called \textbf{Filtered}.
f
The relations can be divided into four categories according to their head and tail entity counts. For any relation, if there is only one head entity in the knowledge graph when given a tail entity, and vice versa, it is an \textbf{1-to-1} relation. If there is only one head entity when given a tail entity but many tail entities when given a head entity, it is an \textbf{1-to-N} relation; Reverse to \textbf{N-to-1} relation; 
If there many head entities when given a tail entity and vice versa, it is an \textbf{N-to-N} relation.

In both Raw and Filtered data, for any type of relations, lower Mean Rank and higher Hits@n indicate the better results and performance.
\vspace{-15pt}
\subsection{Hyper-parameters}
When training the baseline models, we choose the best-selected parameters presented in their original papers. A Translation-based model must contains five parameters, namely the word vector size \(k\), the learning rate \(\lambda\)\emph{,} the margin \(\gamma\), the batch size \(B\) and dissimilarity measure \(d\). 
  
\begin{table}[t]
  \centering \small
  \label{table:statistics}
  \setlength{\tabcolsep}{0.9mm} 
  \begin{tabular}{|c|c|c|c|c|c|c|} 
  \hline
  \bf{Dataset} &\bf{\#Ent}& \bf{\#Rel}& \bf{\#Train}& \bf{\#Valid}& \bf{\# Test}& \bf{\# Domain}\\
  \hline
  WN18 & 40,943 & 18 & 141,442 & 5,000 & 5,000 & 36\\
  \hline
  FB15K & 14,951 & 1,345 & 483,142 & 50,000 & 59,071 & 2690\\
  \hline
  \end{tabular}
  \caption{\#Ent is the entity number of the database, \#Rel is the relation number of the database. \#Train/\#Valid/\#Test are the triples in training/validation/test sets.}
  \vspace*{-15pt}
  \end{table}

\begin{table}[t]
  \centering
  \small
  \setlength{\tabcolsep}{1mm} 
  \label{label:link_prediction_rel}
  \begin{tabular}{|c|c|cc|cc|}
  \hline
  \multirow{3}{*}{\textbf{Metric}} & \multirow{3}{*}{\tabincell{c}{\textbf{external}\\\textbf{Information}}} & \multicolumn{2}{|c|}{\textbf{WN18}} & \multicolumn{2}{|c|}{\textbf{FB15K}} \\
                          & & \textbf{Mean Rank}  &\textbf{Hits@10} & \textbf{Mean Rank}& \textbf{Hits@10} \\ 
                          & & \textbf{Filtered} &\textbf{Filtered}&\textbf{Filtered}&\textbf{Filtered}\\
  \hline
  TransE \cite{transE}&\multirow{12}{*}{NA} &251&89.2&125&47.1\\
  TransH \cite{transH}& &388&82.3&87&64.4\\
  TransR \cite{TransR}& &225&92.0&77&68.7\\
  TranSparse \cite{TranSparse}& &221&93.9&82&79.9\\
  STransE \cite{STransE}& &206&93.4&69&79.7\\
  TransA \cite{TransA}& &\textbf{153}&-&58&-\\
  ITransF \cite{ItransF}& &205&95.2&65&81.4\\
  \textbf{DRE(TransE)}&\multirow{3}{*} &201&93.3&60&74.5 \\
  \textbf{DRE(TransR)}& &165&96.5&38&82.1 \\
  \textbf{DRE(STransE)}& &154&\textbf{96.9}&\textbf{36}&\textbf{83.4} \\
  \hline
 
  \end{tabular}
  \caption{Link prediction results on two datasets.}
  \vspace*{-20pt}
  \end{table}

  \begin{itemize}

    \item[-]For TransE, the configurations are:
    \(k = 50,\ \ \lambda = 0.001,\ \gamma = 2,\text{\ B} = 120\), and
    \(d = L_{1}\) on WN18;
    \(k = 50,\ \ \lambda = 0.001,\ \gamma = 1,\ B = 120\ \)and \(d = L_{1}\)
    on FB15K.
    \item[-]For TransR,
    the configurations are:
    \(k = 50,\ \ \lambda = 0.001,\ \gamma = 4,\ B = 1440\), and
    \(d = L_{1}\) on WN18;
    \(k = 50,\ \ \lambda = 0.001,\ \gamma = 1,\ B = 4800\ \)and
    \(d = L_{1}\) on FB15K.
    \item[-]For STransE, the configurations are:
    \(k = 50,\ \ \lambda = 0.0005,\ \gamma = 5,\ B = 120\), and
    \(d = L_{1}\) on WN18;
    \(k = 100,\ \ \lambda = 0.0001,\ \gamma = 1,\ B = 120\ \)and
    \(d = L_{1}\) on FB15K.
 
  \end{itemize}
In our DRE model, we run SGD for 500 epochs to estimate the parameters
of hyper-ellipsoids, with the learning rate
\(\lambda \in \left\{ \ 0.000001; \ 0.00001;\ 0.0001;\ 0.001; \right\}\) and
the batch size \(B = 120\). The vector size \(\text{k\ }\)is set the
same as the baseline model's vector size. We finally chose
\(\lambda = 0.00001\) as our learning rate according to the results. Except for \(\lambda\) and \(B\), our model does not need other hyper-parameters.
\vspace*{-5pt}
  \begin{table}[t]
    \centering
    \label{label:link_prediction_entity}
  \begin{tabular}{|c|cccc|cccc|}
  \hline
  \multirow{2}{*}{\textbf{Model}}& \multicolumn{4}{|c|}{\textbf{WN18}}& \multicolumn{4}{|c|}{\textbf{FB15K}} \\
  & \textbf{MR}&\textbf{H@10}&\textbf{H@3}&\textbf{H@1}&\textbf{MR}&\textbf{H@10}&\textbf{H@3}&\textbf{H@1}\\
  \hline
  TransE  &251&89.2&83.2&34.2&125&47.1&49.4&20.9\\
  \textbf{DRE(TransE)}&204&93.3&86.2&68.9&60&74.5&57.5&41.9\\
  \hline
  TransR  &225&92.0&89.8&48.6&77&68.7&50.4&21.5\\
  \textbf{DRE(TransR)}&164&96.5&94.4&71.1&38&82.1&70.6&55.5\\
  \hline
  STransE &206&93.4&90.2&70.0&69&79.7&56.8&26.5\\
  \textbf{DRE(STransE)} &154&96.9&94.7&75.3&36&83.4&72.4&58.7\\
  
  \hline
  \end{tabular}
    
  \caption{Results on WN18 and FB15K.MR is mean rank, H@10/3/1 are Hits@10/3/1. In the results of TransE, TransR, and STransE, for MR and Hits@10, we use the results reported in the original papers; For Hits@3/1, we use our own results.}
  \vspace*{-15pt}
  \end{table}

  \vspace{-10pt}
\subsection{Results}

  Table 2 shows the link prediction results of previous work and our method by the Mean Rank and Hit@10. The first 12 rows are prior models without external information; the next three rows are our model DRE with three baseline models,namely DRE(TransE), DRE(TransR) and DRE(STransE), respectively. Our model is also external information free.  Models using external information can achieve better results compared with external-information-free models, but this is not a direct comparison.

  Among models without any external information, our model has achieved best results in both mean rank and Hits@10 on WN18 and FB15K, for example, our model improves Hits@10 of STransE from 93.4\%/79.7\% to 96.9\%/83.4\%  and lower Mean Rank from 206/69 to 154/36 on WN18/FB15K, respectively.

  \textbf{Mean Rank}. For Mean Rank, our model shows strong benefits, significantly reducing extreme cases in link prediction. Taking tail prediction by TransE as an example. In some cases (which are frequent in our test set), \(h + r\) can be very distant from the \(t\) in the hyperspace, which makes the rank of correct triples very high. However, when \(h + r\) is far away from the right entity \(t\), our model gives a large penalty to the incorrect entities around \(h + r\), while only giving a small or none penalty to the correct entities, which are not away from the domain. In this way, the extreme cases are reduced significantly.
  

  \begin{table*}[t]
    \centering
      \label{label:mapping_property}
      \setlength{\tabcolsep}{1mm}
    \begin{tabular}{|c|cccc|cccc|}
    \hline
    \multirow{3}{*}{\textbf{Model}} &\multicolumn{4}{|c|}{\textbf{Predicting Head (Hits@10)}}&\multicolumn{4}{|c|}{\textbf{Predicting Tail (Hits@10)}}\\
      & \textbf{1-to-1} & \textbf{1-to-N } & \textbf{N-to-1} & \textbf{N-to-N} & \textbf{1-to-1} & \textbf{1-to-N 8.9\%} & \textbf{N-to-1 } & \textbf{N-to-N} \\
      & \textbf{1.4\%} & \textbf{8.9\%} & \textbf{14.7\%} & \textbf{75.0\%} & \textbf{1.4\%} & \textbf{8.9\%} & \textbf{14.7\%} & \textbf{75.0\%}\\
    \hline
    TransE & 43.7& 65.7 & 18.2 & 47.2 & 43.7 & 19.7 & 66.7 & 50.0\\
    \textbf{DRE(TransE)} & \textbf{79.6}&\textbf{85.1}&\textbf{47.1}&\textbf{75.6}&\textbf{79.8}&\textbf{52.8}&\textbf{84.1}&\textbf{78.0}\\
    \hline
    TransR & 78.8&89.2&34.1&69.2&79.2&38.4&90.4&72.1\\
    \textbf{DRE(TransR)} & \textbf{86.2}&87.8&\textbf{63.1}&\textbf{81.5}&\textbf{88.1}&\textbf{71.0}&89.0&\textbf{85.7}\\
    \hline
    STransE &82.8&94.2&50.4&80.1&82.4&56.9&93.4&83.1\\
    \textbf{DRE(STransE)} &\textbf{87.1}&88.9&\textbf{69.2}&\textbf{82.2}&\textbf{89.1}&\textbf{74.9}&89.2&\textbf{86.5}\\
    \hline
    \end{tabular}
  
    \caption{Experimental results on FB15K by mapping properties of relations.} 
    \vspace*{-20pt}
    \end{table*}
  
    \vspace{-15pt}
  \subsection{Discussion}
  \textbf{Table 3} shows that our model gives improvements on every evaluation  metric over the baseline models. Also, the results of DRE-(TransE) are roughly similar to the STransE results. Similar to STransE to some extent,  we model the head and tail entities for each relation, respectively. Our model represents its head domain and tail domain, while STransE creates a head projection matrix and a tail projection matrix. STransE/TransR describe domain information by using projection matrices. However, their domain information is implicit, and difficult to ally beyond their models. Besides, they aim to separate entities in the same domain from each other. In contrast, our model aims to extract common knowledge of one domain, using it to separate entities in the domain from entities out of the domain. Notably, our model also improves Hits@1 over the baseline models significantly, especially on FB15K.

  \textbf{Hits@10}. In Table 4, we analyze why our model works on Hits@10. On dataset FB15K, the testing triples with ``1-to-1'' relation take up only 1.4\% of the dataset while the ratio is 8.9\% for ``1-to-N'' relation, 14.7\% for ``N-to-1'' relation and 75.0\% for ``N-to-N'' relation. As a result, instances on ``to-N'' are much more than those on ``to-1''. Our model significantly boosts the performance of predicting ``to-N'' cases. In predicting head entities for ``N-to-1'' relations and tail entities for ``1-to-N'' relations, where the results in baseline models are relatively lower. For ``N-to-N'' relations, our model also improves over the baseline significantly. Since ``to-N'' domains account for 86.8\% (8.9\%/2+14.7\%/2+75.0\%) of testing instances, the overall results were also greatly enhanced.
  
  For some cases in predicting ``to-1'', our model gives lower accuracies. It may be because lack of training data for those domains. For example, for ``1-to-N'' relations, the head entities are much fewer than the tail entities, which leads to lack of entities in head domains. As a result, the domains may not be well trained. The same occurs to the tail domains of ``N-to-1'' relations. But for ``to-N'' domains, the training entities are much more, and the ellipsoids are better trained to represent the domains.

  The baseline models are weak in predicting the head of ``N-to-1'' and predicting the tail of ``1-to-N'' relations, which is the major source of errors. Our method addresses such weakness.

  \vspace{-5pt}
  \section{Conclusion}
  \vspace{-5pt}
  We have shown that a conceptually simple domain model is effective for enhancing the embeddings of knowledge graph by offering hierarchical knowledge. In addition, hyper-ellipsoids are used to represent domains in the vector space, and the distance between an entity and certain domain is used to infer whether the entity belongs to the domain and how much discrepancy they have. Our model can be used over various other KG embedding models to help improve their performance. Results on link prediction show that our model significantly improves the accuracies of state-of-the-art baseline knowledge graph embeddings.  

  To our knowledge, we are the first to learn explicit hierarchical knowledge structure over knowledge graph embeddings. Future work includes extending our model to other related NLP problems, such as information extraction.

\footnotesize
\bibliography{nlpcc}
\bibliographystyle{splncs04}
\end{document}